\documentclass[11pt,a4paper]{article}
\usepackage[hyperref]{acl2019}
\usepackage{times}
\usepackage{latexsym}
\usepackage{url}

\PassOptionsToPackage{hyphens}{url}
\usepackage{hyperref}
\usepackage[hyphenbreaks]{breakurl}
\usepackage[german,english]{babel}

\usepackage[utf8]{inputenc}
\usepackage{graphicx}

\usepackage{caption}
\usepackage{subcaption}

\usepackage{amsmath}

\usepackage{multirow}

\hypersetup{draft}

\title{TwistBytes - Hierarchical Classification at GermEval 2019: walking the fine line (of recall and precision)}

\aclfinalcopy 

\author{Fernando Benites \\
  \texttt{ benf@zhaw.ch} \\[0.4cm] 
   Zurich University of Applied Sciences,\\  Switzerland\\
  }

\date{}

\begin{document}
\maketitle
\begin{abstract}
We present here our approach to the GermEval 2019 Task 1 - Shared Task on hierarchical classification of German blurbs. We achieved first place in the hierarchical subtask B and second place on the root node, flat classification subtask A. In subtask A, we applied a simple multi-feature TF-IDF extraction method using different n-gram range and stopword removal, on each feature extraction module. The classifier on top was a standard linear SVM. For the hierarchical classification, we used a local approach, which was more light-weighted but was similar to the one used in subtask A. The key point of our approach was the application of a post-processing to cope with the multi-label aspect of the task, increasing the recall but not surpassing the precision measure score. 
\end{abstract}

\section{Introduction}

Hierarchical Multi-label Classification (HMC) is an important task in Natural Language Processing (NLP). Several NLP problems can be formulated in this way, such as  patent, news articles, books and  movie genres classification (as well as many other classification tasks like diseases, gene function prediction). Also, many tasks can be formulated as hierarchical problem in order to cope with a large amount of labels to assign to the sample, in a divide and conquer manner (with pseudo meta-labels).
A theoretical survey exists \cite{silla2011survey} discussing on how the task can be engaged, several approaches and the prediction quality measures. Basically, the task in HMC is to assign a sample to one or many nodes of a Directed Acyclic Graph (DAG) (in special cases a tree) based on features extracted from the sample. In the case of possible multiple parent, the evaluation of the prediction  complicates heavily, for once since several paths can be taken, but only in a joining node must be considered.

The GermEval 2019 Task 1 - Shared Task on hierarchical classification of German blurbs focus on the concrete challenge of classifying short descriptive texts of books into the root nodes (subtask A) or into the entire hierarchy (subtask B). The hierarchy can be described as a tree and consisted of 343 nodes, in which there are 8 root nodes. With about 21k samples it was not clear if deep learning methods or traditional NLP methods would perform better. Especially, in the subtask A, since for subtask B some classes had only a few examples. Although an ensemble of traditional and deep learning methods could profit in this area, it is difficult to design good heterogeneous ensembles.

Our approach was a traditional NLP one, since we employed them successfully in several projects \cite{BenitesdeAzevedoeSouza2017Multi,benites2017ST,benites2019twistbytes}, with even more samples and larger hierarchies. We compared also new libraries and our own implementation, but focused on the post-processing of the multi-labels, since this aspect seemed to be the most promising improvement to our matured toolkit for this task. This means but also, to push recall up and hope to not overshot much over precision.

\section{Related Work}

The dataset released by \cite{lewis2004ran} enabled a major boost in
HMC on text. This was a seminating dataset since not only was very large (800k documents)  but the hierarchies were large (103 and 364). Many different versions were used in thousands of papers. Further, the label density \cite{Tsoumakas07} was considerably high allowing also to be treated as multi-label, but not too high as to be disregarded as a common real-world task. Some other datasets were also proposed (\cite{partalas2015lshtc}, \cite{MenciaF10}), which were far more difficult to classify. This means consequently that  a larger mature and varied collection of methods were developed, from which we cannot cover much in this paper.

An overview of hierarchical classification was given in \cite{silla2011survey} covering many aspects of the challenge. Especially, there are local approaches which focus on only part of the hierarchy  when classifying in contrast to the global (big bang) approaches. 

A difficult to answer question is about which hierarchical quality prediction measure to use since there are dozens of. An overview with a specific problem is given in \cite{brucker2011empirical}. An approach which was usually taken was to select several measures, and use a vote, although many measures inspect the same aspect and therefore correlate, creating a bias. The GermEval competition did not take that into account and concentrates only on the flat micro F-1 measures\footnote{The harmonic mean between micro recall and precision gives more weight for the predominantely label. Many new tasks consider the macro averaged F-1 since it gives equal weights for all labels which can be interesting for a large amount of labels (or samples to come).}.

Still, a less considered problem in HMC is the number of predicted labels, especially regarding the post-processing of the predictions\footnote{This is especially important if macro F-1 is used as quality prediction measure, in order to predict as many labels as possible.}.
We discussed this thoroughly in 
\cite{BenitesdeAzevedoeSouza2017Multi}. The main two promising approaches were proposed by \cite{Yang99} and \cite{read2009classifier}. The former focuses on column and row based methods for estimating the appropriate threshold to convert a prediction confidence into a label prediction. \cite{read2009classifier} used the label cardinality (\cite{Tsoumakas07}), which is the mean average label per sample, of the training set and change the threshold globally so that the test set achieved similar label cardinality.

\section{Data and Methodology}

\subsection{Task Definition and Data Description}

The shared task aimed at 
Hierarchical Multi-label Classification (HMC) of Blurbs.  
Blurbs are short texts consisting of some  German sentences. Therefore, a standard framework of word vectorization can be applied. There were 14548 training, 2079 development, and 4157 test samples.

The hierarchy can be considered as an ontology, but for the sake of simplicity, we regard it as a simple tree, each child node having only on single parent node, with 4 levels of depth, 343 labels from which 8 are root nodes, namely: 'Literatur \& Unterhaltung', 'Ratgeber', 'Kinderbuch \& Jugendbuch', 'Sachbuch', 'Ganzheitliches Bewusstsein', 'Glaube \& Ethik', and 'Künste, Architektur \& Garten'.

The label cardinality of the training dataset was about 1.070 (train: 1.069, dev: 1.072) in the root nodes, pointing to a clearly low multi-label problem, although there were samples with up to 4 root nodes assigned. This means that the traditional machine learning systems would promote single label predictions.
Subtask B has a label cardinality  of 3.107 (train: 3.106, dev: 3.114), with 1 up to 14 labels assigned per sample. Table \ref{tab:td-ns} shows a short dataset summary by task.

\begin{table}[ht]
  \centering
  \begin{tabular}{|ccccc|}
    \hline
    Task& samples& labels & cardinality &  density\\\hline
    subtask A&  20,784&8&  1.069& 0.1336\\
    subtask B& 20,784&343& 3.11& 0.0091\\\hline
  \end{tabular}
\caption{Specs for dataset for subtasks A and B }
\label{tab:td-ns}
\end{table}

\subsection{System Definition}

We used two different approaches for each subtask. In subtask A, we used a heavier feature extraction method and a linear Support-Vector-Machine (SVM) whereas for subtask B we used a more light-weighted feature extraction with same SVM but in a local hierarchical classification fashion, i.e. for each parent node such a base classifier was used. We describe in the following the approaches in detail. 
They were designed to be light and fast, to work almost out of the box, and to easily generalise.

\subsubsection{Classifiers}\label{sec:gdi_clf}
\paragraph{Base Classifier} 

\begin{figure}
\centering
{\includegraphics[width=0.5\textwidth]{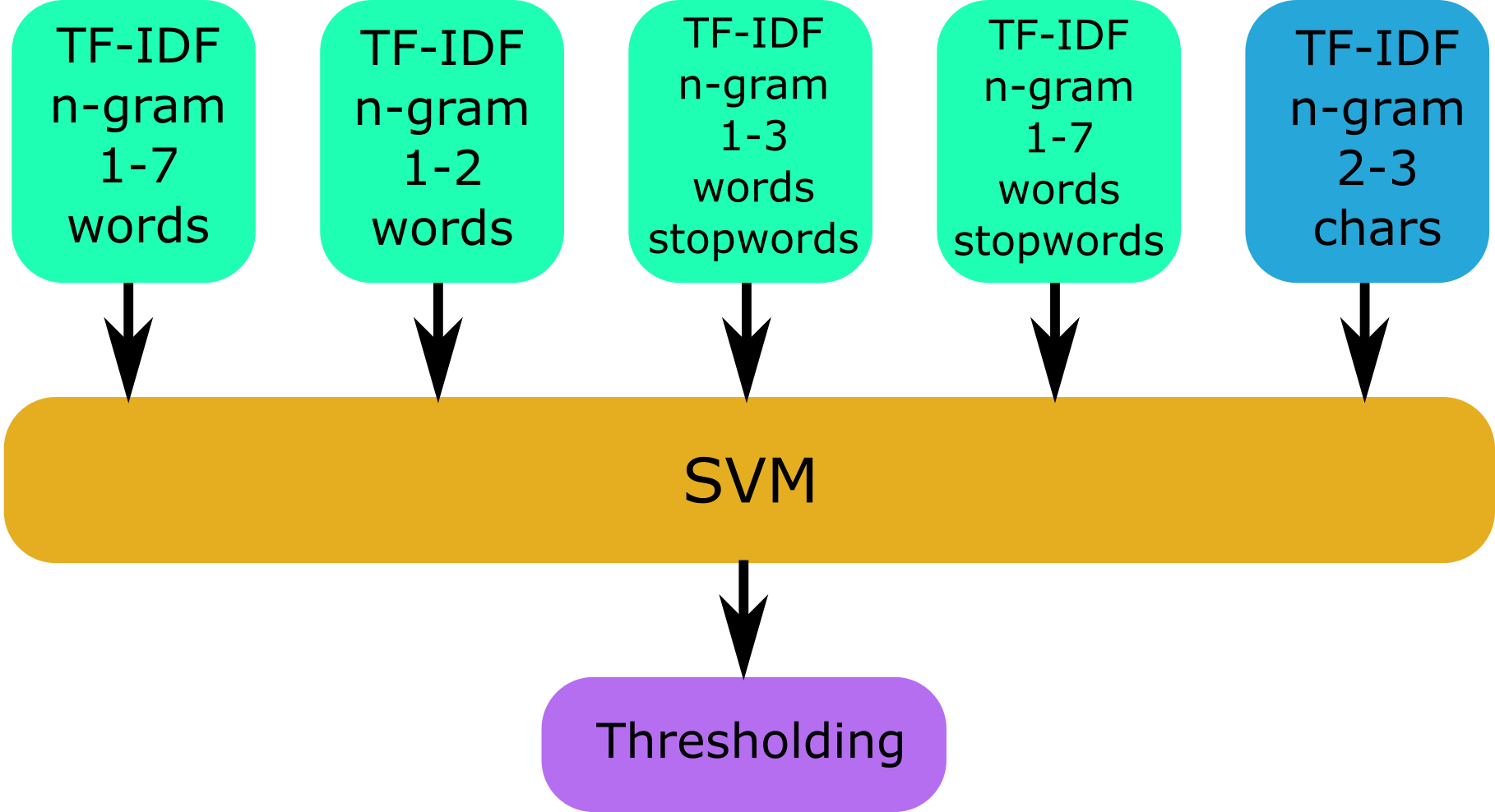}}
\caption{SVM-TF-IDF classifier with ensemble of textual features}
\label{fig:mc}
\end{figure}

For subtask A, we use the one depicted in Fig. \ref{fig:mc}, for subtask B, a similar more light-weight approach was employed as base classifier (described later). As can be seen,  several vectorizer based on different n-grams (word and character) with a maximum of 100k features and preprocessing, such as using or not stopwords, were applied to the blurbs. The term frequency obtained were then weighted with inverse document frequency (TF-IDF). The results of five different feature extraction and weighting modules were given as input for a vanilla SVM classifier  (parameter C=1.5) which was trained in a one-versus-all fashion. 

\subsubsection{Hierarchical Classifier}

We use a local parent node strategy, which means the parent node decides which of its children will be assigned to the sample. This creates also the necessity of a virtual root node. For each node the same base classifier is trained independently of the other nodes. We also adapt each feature extraction with the classifier in each single node much like \cite{paes2014exploring}.
As base classifier, a similar one to Fig. \ref{fig:mc} was used, where only one 1-7 word n-gram, one 1-3 word n-gram with German stopwords removal and one char 2-3 n-gram feature extraction were employed, all with maximum 70k features. 
We used two implementations achieving very similar results. In the following, we give a description of both approaches. 

\paragraph{Recursive Grid Search Parent Node}
Our implementation is light-weighted and optimized for a short pipeline, however for large amount of data, saving each local parent node model to the disk. However, it does not conforms the way scikit-learn is designed. Further, in contrast to the Scikit Learn Hierarchical, we give the possibility to optimize with a grid search each feature extraction and classifier per node. This can be quite time consuming, but can also be heavily parallelized. In the final phase of the competition, we did not employ it because of time constrains\footnote{The system was trained on a Intel Xeon 32 cores and 100 Gb RAM.} and  the amount of experiments performed in the Experiments Section was only possible with a light-weighted implementation.

\paragraph{Scikit Learn Hierarchical}
Scikit Learn Hierarchical\footnote{https://github.com/globality-corp/sklearn-hierarchical-classification/} (Hsklearn)  was forked and improved to deal with multi-labels for the task\footnote{\url{https://github.com/fbenites/sklearn-hierarchical-classification/}}, especially, allowing each node to perform its own preprocessing. This guaranteed that the performance of our own implementation was surpassed and that a contribution for the community was made. This ensured as well that the results are easily reproducible.

\subsubsection{Post-processing: Threshold}

Many classifiers can predict a score or confidence about the prediction. Turning this score into the prediction is usually performed by setting a threshold, such as 0 and 0.5, so labels which have a score assigned greater than that are assigned to the sample. 
This might be  not the optimal threshold in the multi-label classification setup and there are many approaches to set it (\cite{Yang99}). Although these methods concentrate in the sample or label, we have had good results with a much more general approach. 

As described in \cite{BenitesdeAzevedoeSouza2017Multi}, 
Read and Pfahringer \cite{read2009classifier} introduce a method (referred hereinafter to as LCA) to
estimate the threshold globally. Their method chooses the threshold that minimizes the
difference between the label cardinality of the training set and the predicted set.
$$t = \underset{t \in [0,1]}{argmin} |LCard(D_T ) - LCard(H_t(D_S))| $$
where $LCard(D_T)$ denotes the label cardinality of training set and $LCard(H_t(D_S))$
the label cardinality of the predictions on test set if $t$ was applied as the threshold. For that the predictions need to be normalized to unity\footnote{Although a sample wise normalization can be applied, we used a normalization over all predicted samples.}. We also tested this method not for the label cardinality over all samples and labels but only labelwise. In our implementation, the scores of the SVM were not normalized, which produced slightly different results from a normalized approach.

For the HMC subtask B, we used a simple threshold based on the results obtained for LCA. Especially, using multiple models per node could cause a different scaling.

\subsection{Alternative approaches}

We also experimented with other different approaches. The results of the first two were left out (they did not perform better), for the sake of conciseness.
\begin{itemize}
    \item Meta Crossvalidation Classifier: \cite{benites2019twistbytes}
    \item Semi-Supervised Learning:  \cite{jauhiainen2018heli, benites2019twistbytes}
    \item Flair: Flair \cite{akbik2018coling} with different embeddings (BERT (out of memory)\footnote{The system was trained in a Intel icore 7 CPU with 32 Gb RAM with a NVIDIA GeForce 1060 6Gb GPU.}, Flair embeddings  (forward and backward German)). Such sophisticated language models require much more computational power and many examples per label.
This was the case for the subtask A but subtask B was not.
\end{itemize}

\section{Experiments}
\label{sec:results}

We divide this Section in two parts, in first we conduct experiments on the development set and in the second on the test set, discussing there the competition results.

\subsection{Preliminary Experiments on Development Set}

The experiments with alternative approaches, such as Flair, meta-classifier and  semi-supervised learning\footnote{The training, dev and test set seems to come from the same distribution, so the  quality prediction of using a semi-supervised method was worse than without.} yielded discouraging results, so we will concentrate in the SVM-TF-IDF  methods. Especially, semi-supervised proved in other setups very valuable, here it worsened the prediction quality, so we could assume the same "distribution" of samples were in the training and development set (and so we concluded in the test set).

In Table \ref{tab:results-dev-a}, the results of various steps towards the final model can be seen. An SVM-TF-IDF model with word unigram already performed very well. Adding more n-grams did not improve, on the contrary using n-grams 1-7 decreased the performance. Only when removing stopwords it improved again, but then substantially. Nonetheless, a character  2-3 n-gram performed best between these simple models. This is interesting, since this points much more to not which words were used, but more on the phonetics\footnote{For the sake of conciseness, we will not discuss it here.}.

Using the ensemble feature model produced the best results without post-processing. The simple use of a low threshold yielded also astonishingly good results. This indicates that the SVM's score production was very good, yet the threshold 0 was too cautious.

\begin{table*}[t]
\center
\begin{tabular}{|c|l|c|}
\hline
\bf Nr.& \bf  Method&	\bf micro F-1  \\
\hline
1&SVM-TF-IDF, word unigram&0.7965 \\
2&SVM-TF-IDF, word unigram, t=-0.25&0.8234 \\
3&SVM-TF-IDF, word n-gram (1-7)&  0.7875\\
4&SVM-TF-IDF, word n-gram (1-7), t=-0.25&0.8152\\
5&SVM-TF-IDF, word n-gram (1-3), stopwords& 0.8075\\
6&SVM-TF-IDF, word n-gram (1-3), stopwords, t=-0.25 & 0.8240 \\
7&SVM-TF-IDF, char n-gram (2-3)& 0.8205 \\
8&SVM-TF-IDF, char n-gram (2-3), t=-0.25& 0.8332\\\hline
9&SVM-TF-IDF, feat. ensemble& 0.8414\\
10&SVM-TF-IDF, feat. ensemble, threshold LCA& 0.8545 \\
10&SVM-TF-IDF, feat. ensemble, threshold LCA normed & 0.8534  \\
11&SVM-TF-IDF, feat. ensemble, threshold LCA-labelwise& \bf  0.8603\\
12&SVM-TF-IDF, feat. ensemble, threshold -0.25&0.8540 \\
13&SVM-TF-IDF, feat. ensemble, threshold -0.2& \bf 0.8557 \\\hline
14&Flair Embeddings German (forward,backward), 60 epochs& 0.8151 \\
15&SVM-TF-IDF, feat. ensemble, threshold LCA, fixing null& \bf 0.8577\\
16&SVM-TF-IDF, feat. ensemble, threshold LCA-labelwise, fixing null& \bf  0.8623\\
\hline
\end{tabular}
\caption{Micro F-1 scores of different approaches on the development set, best four values marked in bold}
\label{tab:results-dev-a}
\end{table*}

\begin{figure}[tp]
   \hskip-0.5cm
 \includegraphics[width=9cm]{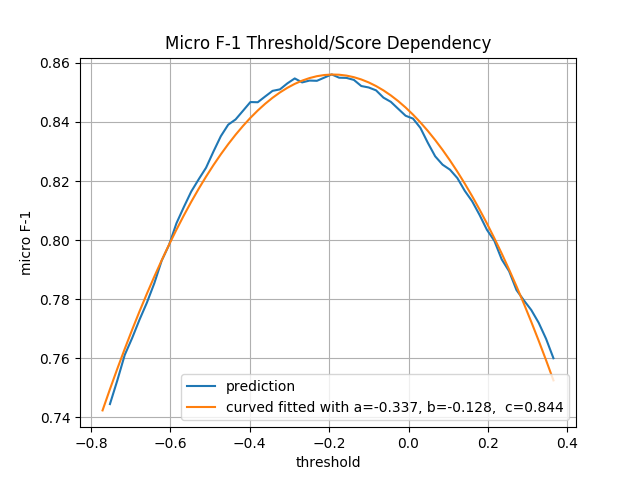}
   \caption{Threshold/micro F-1 dependency}
   \label{fig:threshold}

\end{figure}

In Fig. \ref{fig:threshold}, a graph showing the dependency between the threshold set and the micro F-1 score achieved in the development set is depicted. The curve fitted was $a*x^2+b*x+c$ which has the maximum at approx. -0.2. We chose -0.25 in the expectation that the test set would not be exactly as the development set and based on our previous experience with other multi-label datasets (such as the RCv1-v2) which have an optimal threshold at -0.3. Also as we will see, the results proved us right achieving the best recall, yet not surpassing the precision score. This is a crucial aspect of the F-1 measure, as it is the harmonic mean it will push stronger and not linearly the result towards the lower end, so if decreasing the threshold, increases the recall linearly and decreases also the precision linearly, balancing both will consequently yield a better F-1 score.

Although in Fig. \ref{fig:threshold}, the curve fitted is parabolic, in the interval between -0.2 and 0, the score is almost linear (and strongly monotone decreasing) giving a good indication that at least -0.2 should be a good threshold to produce a higher F-1 score without any loss.

Even with such a low threshold as -0.25, there were samples without any prediction. We did not assign any labels to them, as such post-process could be hurtful in the test set, although in the development it yielded the best result (fixing null).

\begin{table*}[t]
\center
\footnotesize
\begin{tabular}{|l|c|c|c|c||c|c|c|c|c|c|c|c|}
\hline
\multirow{2}{*}{\bf Label} & \multicolumn{4}{c||}{t=0}         &             \multicolumn{4}{c|}{t=-0.25} &\multicolumn{4}{c|}{LCA}                  \\\cline{2-13}
&tn&fp&fn&tp&tn&fp&fn&tp&tn&fp&fn&tp\\\hline\hline
Architektur \& Garten&2062&0&4&13&2061&1&2&15&2061&1&2&15\\
Ganzheitliches Bewusstsein&1959&8&45&67&1951&16&29&83&1951&16&30&82\\
Glaube \& Ethik&1986&3&31&59&1983&6&23&67&1984&5&24&66\\
Kinderbuch \& Jugendbuch&1783&8&80&208&1759&32&50&238&1762&29&51&237\\
Künste&2061&0&6&12&2061&0&4&14&2061&0&4&14\\
Literatur \& Unterhaltung&874&98&58&1049&801&171&31&1076&807&165&31&1076\\
Ratgeber&1799&20&110&150&1781&38&75&185&1785&34&77&183\\
Sachbuch&1701&40&148&190&1672&69&106&232&1674&67&111&227\\\hline
Total&14225&177&482&1748&14069&333&320&1910&14085&317&330&1900\\\hline
\hline
\end{tabular}
\caption{Confusion matrix between label and others for threshold (t) =0 and =-0.25 (true negative: tp, false negative: fn, false positive: fp, true positive: tp) }
\label{tab:results-confsubtaska}
\end{table*}

In Table \ref{tab:results-confsubtaska}, the results of the one-vs-all approach regarding the true negative, false positives, false negatives and true positives for the different threshold 0, -0.25 and LCA are shown. Applying other threshold than 0 caused the number of true positives to increase without much hurting the number of true negatives. In fact, the number of false positives and false negatives became much more similar for -0.25 and LCA than for 0. This results in the score of recall and precision being also similar, in a way that the micro F-1 is increased. Also, the threshold -0.25 resulted that the number of false positive is greater than the number of false negatives, than for example -0.2. LCA produced similar results, but was more conservative having a lower false positive and higher true negative and false negative score.

We also noticed that the results produced by subtask A were better than that of subtask B for the root nodes, so that a possible crossover between the methods (flat and hierarchical) would be better, however we did not have the time to implement it. Although having a heavier feature extraction for the root nodes could also perform similar (and decreasing complexity for lower nodes).
We use a more simple model for the subtask B so that it would be more difficult to overfit.

\begin{table}[t]
\center
\begin{tabular}{|l|c|}
\hline
\bf Method&	\bf micro F-1\\
\hline
Hsklearn & 0.6544\\
Hsklearn, t=-0.25 &\bf 0.6758\\
Hsklearn, t=-0.2 &  0.6749\\
Hsklearn, LCA normalized & 0.6645 \\
Hsklearn, LCA & 0.6717 \\
Hsklearn extended & 0.6589\\
Hsklearn extended, t=-0.25 &\bf 0.6750\\
Hsklearn extended, t=-0.2 & \bf 0.6765\\
own imp. & 0.6541\\
own imp., t=-0.25 & 0.6704\\
own imp., t=-0.2 & 0.6715\\
\hline
\end{tabular}
\caption{Preliminary experiments on subtask B, best three values marked in bold}
\label{tab:dev-b}
\end{table}

Table \ref{tab:dev-b} shows the comparison of the different examined approaches in subtask B in the preliminary phase. 
Both implementations, Hsklearn and our own produced very similar results, so for the sake of reproducibility, we chose to continue with Hsklearn. We can see here, in contrary to the subtask A, that -0.25 achieved for one configuration better results, indicating that -0.2 could be overfitted on subtask A and a value diverging from that could also perform better. The extended approach means that an extra feature extraction module was added (having 3 instead of only 2) with n-gram 1-2 and stopwords removal. The LCA approach yielded here a worse score  in the normalized but almost comparable in the non-normalized. However, the simple threshold approach performed better and therefore more promising.

\subsection{Subtask A}

\begin{table*}[t]
\center
\begin{tabular}{|l|c|c|c|c|}
\hline
\bf Rank&	\bf  Teamname&	\bf precision&	\bf recall&	\bf micro F-1\\
\hline

1&	EricssonResearch&0.8923	&0.8432&	0.8670\\
- &twistbytes LCA fixing null& 0.8536& 0.8790& 0.8661\\
- &twistbytes LCA-labelwise fixing null& 0.8536 & 0.8763 &0.8648\\
2&	twistbytes&0.8650&0.8617	&0.8634\\
3&	DFKI-SLT&0.8760&	0.8472	&0.8614\\
4&	Raghavan&0.8777&	0.8383	&0.8575\\
5&	knowcup&0.8525&	0.8362	&0.8443\\
6&	fosil-hsmw&0.8427&	0.832&	0.8373\\
7&	Averbis&0.8609&	0.8083	&0.8337\\
8&	HSHL1&0.8244&	0.8159	&0.8201\\
9&	Comtravo-DS&0.8144&	0.8255&	0.8199\\
10&	HUIU&0.8063&	0.8072	&0.8067\\
11&	LT-UHH&0.8601&	0.7481	&0.8002\\
\hline
\end{tabular}
\caption{Results of subtask A, best micro F-1 score by team}
\label{tab:results-a}
\end{table*}
In Table \ref{tab:results-a}, the best results by team regarding micro F-1 are shown. Our approach reached second place. The difference between the first four places were mostly  0.005 between each, showing that only a minimal change could lead to a place switching. 
Also depicted are  not null improvements results, i.e. in a following post-processing, starting from the predictions, the highest score label is predicted for each sample, even though the score was too low.
It is worth-noting that the all but our approaches had much higher precision compared to the achieved recall. 

Despite the very high the scores, it will be difficult to achieve even higher scores with simple NLP scores. Especially, the n-gram TF-IDF with SVM could not resolve descriptions which are science fiction, but are written as non-fiction book\footnote{Exemplary are books describing dystopias which from a n-gram perspective have very much the same vocabulary of a non-fiction book. Here, more aspects of the language need to be captured, such as a focus to constructions like "in a future New York City".}, where context over multiple sentences and word groups are important for the prediction.

\subsection{Subtask B}
\begin{table*}[t]
\center
\begin{tabular}{|l|c|c|c|c|}
\hline
\bf Rank&	\bf  Teamname&	\bf precision&	\bf recall&	\bf micro F-1\\
\hline
1&twistbytes&0.7072&	0.6487&0.6767\\
2&EricssonResearch&0.7377&	0.6174&0.6722\\
3&knowcup&0.7507&	0.5808&0.6549\\
4&Averbis&0.677&	0.614&0.644\\
5&DFKI-SLT&0.7777&	0.5151&0.6197\\
6&HSHL1&0.7216&	0.5375&0.6161\\
7&Comtravo-DS&0.7042&	0.5274&0.6031\\
8&LT-UHH&0.8496&	0.3892&0.5339\\
9&NoTeam&0.4166&	0.276&0.332\\
10&DexieDuo&0.0108&	0.0034&0.0052\\\hline
\end{tabular}
\caption{Results of subtask B, best micro F-1 score by team}
\label{tab:results-b}
\end{table*}

The best results by team of subtask B are depicted in Table \ref{tab:results-b}.
We achieved the highest micro F-1 score and the highest recall. Setting the threshold so low was  still too high for this subtask, so precision was still much higher than recall, even in our approach. We used
many parameters from subtask A, such as C parameter of SVM and threshold. However, the problem is much more complicated and a grid search over the nodes did not complete in time, so many parameters were not optimised. Moreover, although it is paramount to predict the parent nodes right, so that a false prediction path is not chosen, and so causing a domino effect, we did not use all parameters of the classifier of subtask A, despite the fact it could yield better results. It could as well have not generalized so good.

The threshold set to -0.25 shown also to produce better results with micro F-1, in contrast to the simple average between recall and precision. This can be seen also by checking the average value between recall and precision, by checking the sum, our approach produced 0.7072+0.6487 = 1.3559 whereas the second team had 0.7377+0.6174 = 1.3551, so the harmonic mean gave us a more comfortable winning marge.

  


\section{Conclusion}

We achieved first place in the most difficult setting of the shared Task, and second on the "easier" subtask. We achieved the highest recall and this score was still lower as our achieved precision (indicating a good balance). We could reuse much of the work performed in other projects building a solid feature extraction and classification pipeline. We demonstrated the need for post-processing measures and how the traditional methods performed against new methods with this problem.  Further, we improve a hierarchical classification open source library to be easily used in the multi-label setup achieving state-of-the-art performance with a simple implementation.

The high scoring of such traditional and light-weighted methods is an indication that this dataset has not enough amount of data to use deep learning methods. Nonetheless, the amount of such datasets will probably increase, enabling more deep learning methods to perform better.

Many small improvements were not performed, such as elimination of empty predictions and using label names as features. This will be performed in future work.

\section{Acknowledgements}

We thank Mark Cieliebak and Pius von Däniken for the fruitful
discussions. We also thank the organizers of the GermEval 2019 Task 1.


\bibliographystyle{acl_natbib}

\end{document}